\documentclass[sigconf]{acmart} 

\makeatletter
\def\@ACM@checkaffil{
    \if@ACM@instpresent\else
    \ClassWarningNoLine{\@classname}{No institution present for an affiliation}%
    \fi
    \if@ACM@citypresent\else
    \ClassWarningNoLine{\@classname}{No city present for an affiliation}%
    \fi
    \if@ACM@countrypresent\else
        \ClassWarningNoLine{\@classname}{No country present for an affiliation}%
    \fi
}
\makeatother

\usepackage[export]{adjustbox}
\usepackage{mathtools}
\usepackage{enumitem}

\usepackage[normalem]{ulem} 
\usepackage{algorithm}
\usepackage[noend]{algpseudocode} 
\usepackage{algorithmicx}

\usepackage{etoolbox}
\usepackage{tikz}
\usetikzlibrary{tikzmark}
\usetikzlibrary{calc}

\newcommand{\ALGtikzmarkcolor}{black}
\newcommand{\ALGtikzmarkextraindent}{4pt}
\newcommand{\ALGtikzmarkverticaloffsetstart}{-.5ex}
\newcommand{\ALGtikzmarkverticaloffsetend}{-.5ex}
\makeatletter
\newcounter{ALG@tikzmark@tempcnta}

\newcommand\ALG@tikzmark@start{%
	\global\let\ALG@tikzmark@last\ALG@tikzmark@starttext%
	\expandafter\edef\csname ALG@tikzmark@\theALG@nested\endcsname{\theALG@tikzmark@tempcnta}%
	\tikzmark{ALG@tikzmark@start@\csname ALG@tikzmark@\theALG@nested\endcsname}%
	\addtocounter{ALG@tikzmark@tempcnta}{1}%
}

\def\ALG@tikzmark@starttext{start}
\newcommand\ALG@tikzmark@end{%
	\ifx\ALG@tikzmark@last\ALG@tikzmark@starttext
	\else
	\tikzmark{ALG@tikzmark@end@\csname ALG@tikzmark@\theALG@nested\endcsname}%
	\tikz[overlay,remember picture] \draw[\ALGtikzmarkcolor] let \p{S}=($(pic cs:ALG@tikzmark@start@\csname ALG@tikzmark@\theALG@nested\endcsname)+(\ALGtikzmarkextraindent,\ALGtikzmarkverticaloffsetstart)$), \p{E}=($(pic cs:ALG@tikzmark@end@\csname ALG@tikzmark@\theALG@nested\endcsname)+(\ALGtikzmarkextraindent,\ALGtikzmarkverticaloffsetend)$) in (\x{S},\y{S})--(\x{S},\y{E});%
	\fi
	\gdef\ALG@tikzmark@last{end}%
}

\apptocmd{\ALG@beginblock}{\ALG@tikzmark@start}{}{\errmessage{failed to patch}}
\pretocmd{\ALG@endblock}{\ALG@tikzmark@end}{}{\errmessage{failed to patch}}

\usepackage{caption}
\usepackage{cleveref}

\crefformat{section}{\S#2#1#3} 
\crefformat{subsection}{\S#2#1#3}
\crefformat{subsubsection}{\S#2#1#3}

\newcommand{\halfgnn}{\textsc{HalfGNN}}
\newcommand{\gnnone}{\textsc{GNNOne}}

\usepackage{color}
\newcommand{\red}[1]{\textcolor{red}{#1}}
\newcommand{\blue}[1]{\textcolor{blue}{#1}}
\newcommand{\td}{{\bf \color{red} TODO~}}

\usepackage{listings}
\usepackage{xcolor}

\definecolor{codegreen}{rgb}{0,0.6,0}
\definecolor{codegray}{rgb}{0.5,0.5,0.5}
\definecolor{codepurple}{rgb}{0.58,0,0.82}
\definecolor{backcolour}{rgb}{0.95,0.95,0.92}

\lstdefinestyle{mystyle}{
  commentstyle=\color{codegreen},
  keywordstyle=\color{magenta},
  numberstyle=\tiny\color{codegray},
  stringstyle=\color{codepurple},
  basicstyle=\ttfamily\footnotesize,
  breakatwhitespace=false,         
  breaklines=true,                 
  captionpos=b,                    
  keepspaces=true,                 
  numbers=left,                    
  numbersep=5pt,                  
  showspaces=false,                
  showstringspaces=false,
  showtabs=false,                  
  tabsize=2
}
\usepackage{balance}

\usepackage{graphicx}
\usepackage{subcaption}

\begin{document}
\date{}


\title{Optimization of GNN Training Through Half-precision}

\author{Arnab Kanti Tarafder} 
\affiliation{ 
  \institution{William \& Mary}
}
\email{aktarafder@wm.edu}
\author{Yidong Gong} 
\affiliation{ 
  \institution{William \& Mary}
}
\email{ygong07@wm.edu}
\author{Pradeep Kumar} 
  \affiliation{ 
    \institution{William \& Mary}
  }
  \email{pkumar@wm.edu}

\copyrightyear{2025}
\acmYear{2025}
\setcopyright{cc}
\setcctype{by}
\acmConference[HPDC '25]{The 34th International Symposium on
High-Performance Parallel and Distributed Computing}{July 20--23, 2025}{Notre
Dame, IN, USA}
\acmBooktitle{The 34th International Symposium on High-Performance Parallel
and Distributed Computing (HPDC '25), July 20--23, 2025, Notre Dame, IN, USA}
\acmDOI{10.1145/3731545.3731575}
\acmISBN{979-8-4007-1869-4/2025/07}



\begin{abstract}
Recent trends in lower precision, e.g. half-precision floating point, training have shown improved system performance and reduced memory usage for Deep Learning while maintaining accuracy. However, current GNN systems cannot achieve such goals for GNN, as our analyses show that they massively underperform while showing abnormal accuracy when using half-precision. These systems suffer from under-utilization of hardware resources, poor training performance, and value overflow issues due to lowered precision. To mitigate this, we introduce {\halfgnn}, a half-precision based GNN system. 
{\halfgnn} proposes novel techniques: new vector operations for half-precision data types that improve data load and reduction performance, and discretized SpMM that overcomes the value overflow and natively provides workload balancing. Such techniques improve hardware utilization, reduce memory usage, and remove atomic writes. Evaluations show that {\halfgnn} achieves on average of $2.30\times$ speedup in training time over DGL (float-based)
for GAT, GCN, and GIN respectively while achieving similar accuracy, and saving 2.67$\times$ memory.




\end{abstract}



\maketitle

%

\section{Introduction} \label{sec.intro}
Graph neural networks (GNN) have become ever more present in many deep learning applications ~\cite{xu2018powerful, courbariaux2015binaryconnect, han2020graph, krahmer2003graph, wang2021relational, wu2022graph, ying2018graph, zitnik2018modeling, fan2019graph, hamilton2018embedding, zhang2018gaan}. This is because graphs can model different kinds of phenomena across various domains. It has been shown that GNNs offer better performances than other approaches~\cite{hamilton2017inductive, xu2018powerful} for graph datasets.
GNN relies on two key sparse kernels: 
\textit{SpMM} (Sparse matrix-matrix multiplication), and \textit{SDDMM} (Sampled dense-dense matrix multiplication) and their many variants. These two kernels are very different from a \emph{GeMM} (General Matrix Multiplication) kernel due to the unstructured sparsity pattern present in the graph dataset. 


GPU memory and computation capability have increased in recent years. However, it is still limited compared to the requirements of training large models. So, recent DL systems have explored lower precision data types such as \textit{half precision} (16-bit floating point) instead of \textit{float} (32-bit floating point) to utilize GPU resources efficiently. The half data type can theoretically provide faster computation and lower memory consumption than float.
PyTorch~\cite{pytorch2019}, a popular DL framework, has introduced mixed-precision training to integrate half data types.  
DGL~\cite{dgl2019}, a popular GNN library, does provide support for \emph{half}, which internally uses the Cusparse \cite{cusparse} library for the \emph{SpMM} kernel, and in-house SDDMM. However, as we explain next, GNN remains to be explored using \emph{half} precision. 



\begin{figure*}[h!]
\centering
\begin{subfigure}{0.33\textwidth}
    \includegraphics[width=\textwidth]{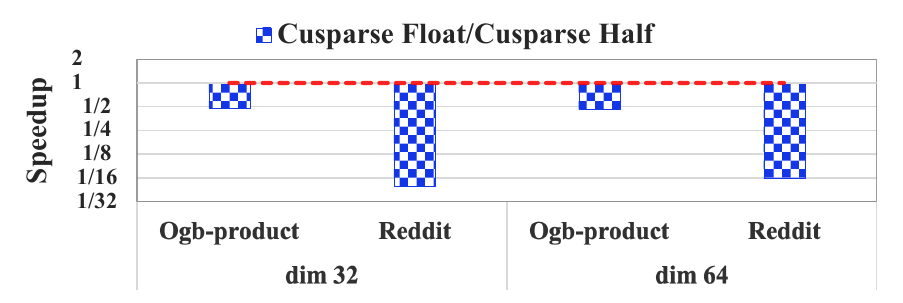}
    \caption{SpMM Runtime Comparison}
    \label{fig-float-half-spmm}
\end{subfigure}\hfill
\begin{subfigure}{0.33\textwidth}
    \includegraphics[width=\textwidth]{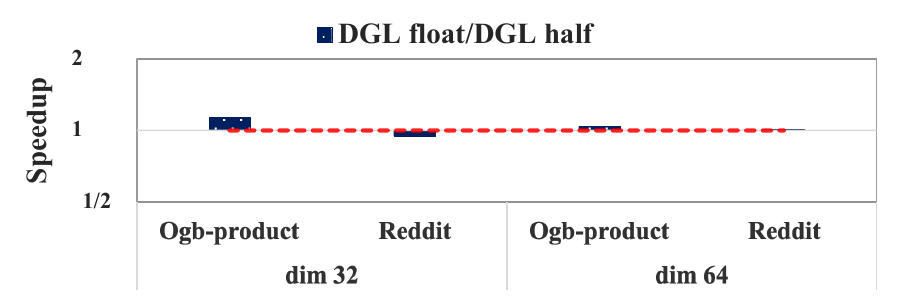}
    \caption{SDDMM Runtime Comparison}
    \label{fig-float-half-sddmmvv}
\end{subfigure} \hfill
\begin{subfigure} {0.33\textwidth}
  \centering
\includegraphics[width=\textwidth,height=0.09\textheight]{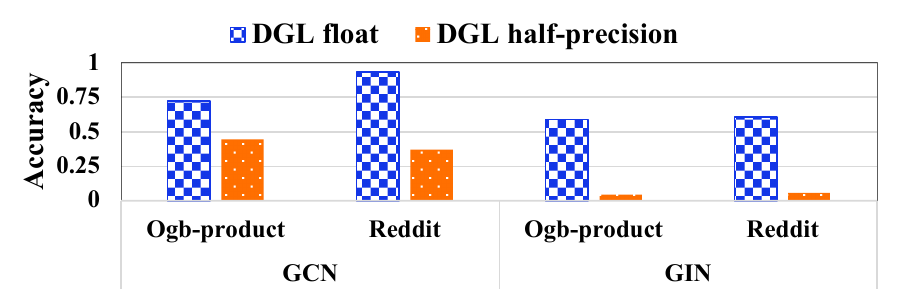} 
  \caption{Training Accuracy Comparison}
  \label{obs-accuracy}
\end{subfigure}
\caption{\small{DGL half-precision Analysis over Float-based DGL: (a) Half-precision \emph{SpMM} is very slow, (b) Half-precision \emph{SDDMM} is similar in runtime than float-based SDDMM, (c) Low accuracy for GCN and GIN in \emph{half-precision} }}
\label{dgl kernel}
\end{figure*}

\noindent \textbf{Training Runtime Efficiency.}
We plotted the SpMM runtime using Cusparse, which DGL uses internally, in Fig.~\ref{fig-float-half-spmm} for different feature lengths on two widely used datasets in GNN literature. The dataset and machine configurations are discussed in ~\cref{sec.exp}. 
The results show that \emph{half-precision} \emph{SpMM} is significantly slower than float SpMM. Similarly, Fig.~\ref{fig-float-half-sddmmvv} shows that DGL \emph{SDDMM} does not improve runtime when the half-precision is used.
\textit{These results indicate that the performance characteristics of sparse kernels for half-precision have not yet been fully explored.}

The performance characteristic is further complicated due to more hardware choices, such as the tensor core-- a hardware unit within the GPU that performs faster computation on \emph{half-precision}. 
To apply tensor cores to sparse kernels, we need to convert the sparse matrix to dense tiles; however, a GNN dataset may not be made up of dense tiles due to its unstructured sparsity of graph datasets.
Unfortunately, TC-GNN~\cite{wang2023tc}, which postulates that tensor cores can provide better SpMM performance, causes huge performance degradation, as shown by numerous prior works~\cite{Graphpy, GNNBench}.






\noindent \textbf{Training Accuracy.}
Most importantly, Fig.~\ref{obs-accuracy} shows that DGL achieves significantly lower training accuracy when using \emph{half-precision} for Ogb-Product and Reddit datasets. 
It would be acceptable if there were small losses in training accuracy due to a loss in precision. Unfortunately, the training loss becomes \emph{NaN}(not a number) after a couple of iterations, questioning \textit{whether the model or the underlying system design is responsible for significantly lower accuracy.}


\noindent \textbf{Contribution.}
These observations motivate us to explore and understand the GNN \textit{accuracy} and \textit{runtime} characteristics when using {half-precision}, resulting in the proposed {\halfgnn}, the first ever half-precision based GNN System that achieves accuracy equivalent to float-based GNN and achieves significant runtime performance. 
We emphasize that the CUDA core is still the best for computation, as the tensor core is unsuited due to the unstructured sparsity in graph datasets. Hence, borrowing the existing idea from float-based sparse kernels becomes relevant. 

Unfortunately, adopting prior float-based techniques fails to bring better performance because of several issues: 
\emph{\textbf{(a)}} data-load does not achieve full memory coalescing in GPU due to the usage of half-precision as each GPU warp can only issue 64 bytes of data-load and not 128 bytes; \emph{\textbf{(b)}} half-based arithmetic can only achieve sub-optimal throughput, thereby lowering the performance; \emph{\textbf{(c)}} current float-based workload balancing techniques handle conflicting writes using atomic instructions, which is very costly for half-precision based operation.

To this end, we propose the following contributions:



\noindent $\bullet$
\textbf{Adopting Half2 Data-Type for a Strong Baseline for Half-Precision Sparse Kernels.}
We adopt \textit{half2} data-type. This native GPU vector data type combines two half-precision numbers for memory operations (data-load and data-store instructions) and computations to achieve full coalesced memory access and better arithmetic throughput. Its adoption faces many challenges that we overcome using \textit{Edge Feature Mirroring}, \textit{Feature Padding for Odd Vertex Features}, and \textit{sub-warps} to impart better thread utilization. 
This establishes an ideal baseline for half-precision kernels, which we analyze, and then propose the following techniques as the baseline fails to provide the envisioned performance and accuracy.

\noindent $\bullet$
\textbf{More Vector Types for Improving SDDMM Performance.}
We group $4$ and $8$ half values, called (\emph{half4} and \emph{half8}) vector data types respectively. The proposed vector data types increase data-load performance by being able to issue more feature load instructions before the implicit memory barrier is encountered in the inter-thread communication for SDDMM reduction. On the other hand, the proposed data types minimize the number of inter-thread communications required in SDDMM.   
\noindent $\bullet$
\textbf{Value Overflow and Discretized Reduction For SpMM.}
We analyze the root cause of the NaN training error in GNNs. A vertex with a large neighborhood size, common in graphs, can produce INF (infinity) due to value overflow in SpMM \textit{reduction} as half-precision has a limited numerical range. This INF then leads to NaN generation due to follow-up operations such as softmax involving arithmetic on two INF.
However, GNN models indeed offer built-in mechanisms to protect against overflow, such as degree-norm scaling in GCN~\cite{kipf2016semi}, which scales the output of SpMM back to half-precision range. 
However, the presence of such a mechanism is insufficient unless explored systematically. 
This work studies many such mechanisms across different GNN models and embeds them into system design through \textit{discretized reduction} where only a finite batch of dot products are reduced and scaled (normalized) before handling the next batch of dot products, thus protecting the output against overflow.  

The discretization is naturally suited to provide workload balancing in SpMM as one is free to choose the batch size, thereby allocating an equal number of computation cores to each discretization unit, leading to workload balancing. 
We demonstrate an edge-parallel and a vertex-parallel SpMM to show the generality of the proposed discretization for workload balancing and propose non-atomic write through usage of \textit{staging buffer} and a follow-up kernel.



\noindent $\bullet$ 
\textbf{Fear of Overflow and Frequent Data-Conversion.}
Our Further analysis shows that a naive mixed-precision system performs excessive data-type conversion of tensors (float to half and vice-versa) or invokes many float-based kernels, thereby not providing much benefit in moving to mixed-precision training.  
For example, the PyTorch exponential operation works in float due to the fear of overflow. This operation is used in the GAT attention mechanism, introducing half to float on its input tensor and float to half on the output tensor. 
However, these conversions are redundant due to the built-in properties of the GAT attention mechanism that guarantee no overflow. This paper studies many such operations to truly incorporate half-precision.

Evaluations show that {\halfgnn} achieves state-of-the-art accuracy along with huge improvement in runtime performance for \emph{half-precision}. Briefly, we reduce the runtime on average by $2.30\times$ and required memory by $2.67\times$ on average over DGL 
for GAT, GCN, and GIN, respectively. Furthermore, {\halfgnn} achieved $7.12\times$, and $22.89\times$ speedup over Cusparse half SpMM, and DGL half SDDMM. 

We applied some of the performance optimizations of {\halfgnn} to Huang et al. ~\cite{huang2021understanding} and achieved $1.79\times$ speedup for SpMM, validating the generality of the performance optimizations. 

The remainder of the paper is organized as follows. We present the necessary background in \cref{sec.background} and analyze the current GNN systems for half-precision, discuss related works, and present an overview in \cref{sec.analysis}. Detailed design is in \cref{sec.half2} and \cref{sec.optimization}, evaluations in \cref{sec.exp}, and conclude in \cref{sec.conclusion}.

\begin{figure*}[t]
  \centering
  \includegraphics[scale= 0.66]{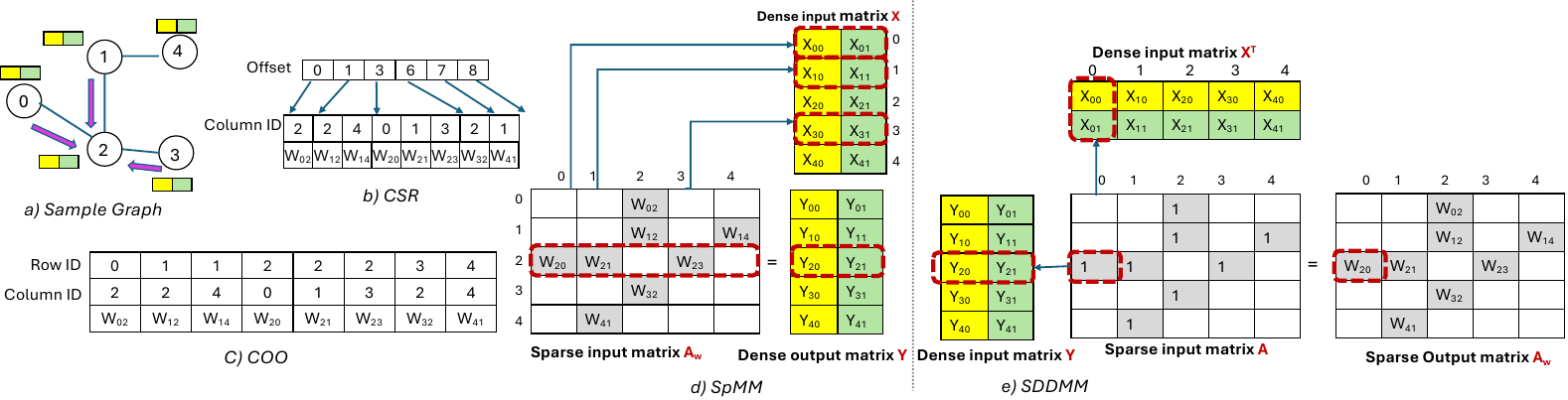} 
 \caption{\small{A sample graph, its storage formats, and illustration of SpMM and SDDMM kernels: the two colors represent different vertex features.}}
 \label{fig-background}
\end{figure*}

\section{Background} \label{sec.background}

A program written for a GPU is called a \textit{kernel}.
Computation units in GPUs are known as \textit{streaming multiprocessors} (SM). SM provides a programmable hardware cache, called \textit{shared memory}, which the kernel can explicitly use to write and read.
Programmers create and configure \textit{Cooperative Thread Arrays} (\textit{CTA}) or \textit{thread-blocks} that are mapped to SM by the GPU driver.
In a CTA, threads are usually grouped into 32 threads, called a \textit{warp}, which executes computation in SIMT (single instruction multiple threads). 
Kernels are generally designed to access memory in a contiguous manner for better data-load performance.

GPU has become a very important accelerator for training diverse deep-learning models. Given this background on GPU, we next describe the graph storage and GNN training steps, explaining the computation steps of major sparse kernels. 

\subsection{Graph Neural Network and Sparse Kernels}  \label{sec.sparse}

\subsubsection{\textbf{Graph Storage Format}}
Fig.~\ref{fig-background} shows a sample graph G = \{V, E\} where V is the vertex set and E is the edge set. We use graph theory and linear algebra terminologies (e.g., row, columns, and non-zero elements (NNZ)) to describe the graph and features of vertices and edges. 
COO (Coordinate Objects) storage format uses two arrays of rowID and ColID, where each $(rowID, colID)$ pair denotes an edge or non-zero element (NZE). The CSR (Compressed Sparse Row) storage format also uses two arrays: \textit{offset array} points to the start of the neighborhood of each vertex or row. The \textit{degree} defines the neighborhood size of each vertex.


We classify tensors involved in kernels as \textit{vertex-level} or \textit{edge-level} tensors. For a graph having $|V|$ vertices and $|E|$ edges, each vertex having $|F|$ features would have a feature matrix of shape $|V|\times|F|$ (vertex-level tensor). We get $|E|\times1$ tensor (edge-level tensor) for edge features.


GNN training, like deep learning,  has two parts: \textbf{forward} computation calls many kernels from the input layer to the output layer, deriving prediction values. They are compared with the ground truth to calculate the loss. \textbf{Backward} computation uses the loss and invokes kernels from the output layer to the input layer to derive gradients and update the model parameters. 
The backward computation accesses \textit{state-tensor}-- results that are produced in the forward computation.

\subsubsection{\textbf{Sparse Kernels}} \label{sec.background.sparse}
The paper deals with these 2 classes of sparse kernels: sparse matrix dense matrix multiplication (SpMM) and sampled dense dense matrix multiplication (SDDMM).
SpMM ($Y \leftarrow A_wX$) is
shown in Fig.~\ref{fig-background}d, where the edges contains features, and hence $A_w$ implies a graph topology ($A$) plus an edge-level tensor ($W_e$). 
It is then multiplied to a vertex-level tensor X of size $|V|\times |F|$, producing another vertex-level tensor $Y$ of size $|V|\times |F|$.

Its backward computation needs to include both the \textit{SpMM} ($\delta X \leftarrow A_w^T \delta Y$) and SDDMM ($\delta W_e \leftarrow A \odot (\delta Y X^T)$), where $\delta$ represents gradients of the tensors. SDDMM is 
shown in Fig.~\ref{fig-background}e produces $\delta W_e$ that represents gradients on edges. 
One can note that X, Y, and $W_e$ are state-tensors.



\emph{SpMM} is categorized into 2 types. 
\textit{SpMMve} is the usual SpMM as described above. 
A special case is \textit{SpMMv} where all edge features are treated as 1.0, hence we do not need to store an explicit $W_e$.
SpMMv is the only sparse kernel that is used in GCN~\cite{kipf2016semi} and GIN~\cite{xu2018powerful}. 
Attention-based GNNs, such as GAT~\cite{gat18iclr}  would require both \emph{SpMMve} and \emph{SDDMM}.


Sparse kernels broadly have \textit{3 key steps}. 
a) \textbf{Data Load} brings graph and features from the GPU memory to the registers of the computation unit. 
b) perform parallel \textbf{dot products} of features: between features of row and column ID for each NZE in case of SDDMM, and between features of NZE (i.e., edge weight) and features of column ID in SpMM.    
c) \textbf{Reduction and Data-Store}  first reduces the computed dot products, followed by writing to the output (GPU) memory: 
SDDMM performs reduction across feature dimensions, so reduction mostly needs inter-thread communication and are stored for each edge; and
SpMM reduction happens across neighborhood dimensions, needing inter-thread communication only when the vertex neighbors are allocated more than one warp to process, as in workload-balanced SpMM and are stored for each row ID.



\subsubsection{\textbf{Parallelism.}} \label{sec.background.parallel}
\textit{Vertex-parallel}~\cite{bell2009implementing} sparse kernels allocate equal computation to process the neighborhood of each vertex. For example, one warp is needed to process the neighborhood of each vertex. Works such as~\cite{gale2020sparse,featgraphsc20, gespmmsc2020, fu2022tlpgnn, wu2021seastar} uses vertex-parallel variants. Workload imbalance occurs when certain vertices have a larger neighborhood than others. 
Edge-parallel systems~\cite{yang2018design,gnnone} divide edges equally among computation, leading to a fully workload-balanced design. However, they introduce the need for inter-thread communication in SpMM for reduction and atomic write for the data-store phase. 

Both systems use \textit{feature-parallel}~\cite{yang2018design} method when loading the vertex-level features: 32 threads of the warp load 32 different features at once, but 2 features are shown in Fig.~\ref{fig-background} for illustration of matrix $X$. This leads to coalesced memory access from $X$.

\subsection{Half-Precision and Vector Data-Type}
\emph{Half-precision} floating point numbers imply 16-bit floating point numbers and hence support a lower range of absolute values: $2^{-14}$  
to $(2-2^{-10}).2^{15}$ i.e. $6.1\times 10^{-5}$ to $6.5\times 10^4$, which is very small compared to the range of single-precision float data-type. Hence, any value greater than the upper bound overflows to INF, while smaller values underflow to zero. Modern GPUs have provided support for half-precision arithmetic.

 \begin{figure}[t]
  \centering
  \includegraphics[scale=0.4]{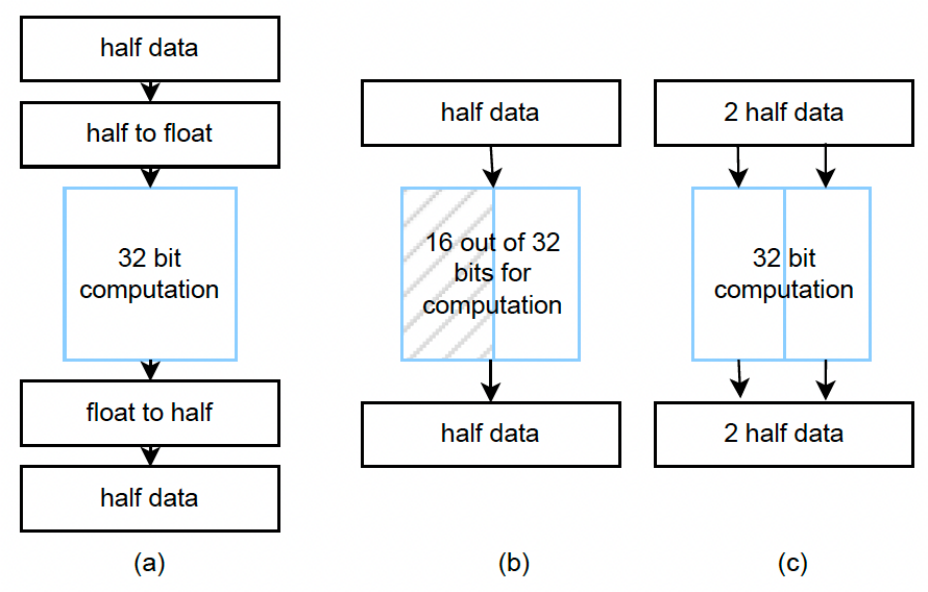} 
  \caption{\small{Arithmetic operations on Half}}
  \label{fig-half-types}
\end{figure}

If one relies on native operators, such as `+' or `*', the data is implicitly converted to float, where the operation is performed in float (Fig.~\ref{fig-half-types}a). 
One can use CUDA intrinsic APIs for half computation that avoids this data-type conversion but achieves the same throughput as the float (Fig.~\ref{fig-half-types}b).

\textit{Vector data types} allow grouping of various lengths of basic types for the data-load and the computation phases. 
For example, two half numbers could be packed in one half2 vector data type, whose size is 32 bits. GPUs support native data-load and arithmetic on the half2 data type (Fig.~\ref{fig-half-types}c), though DGL does not use it in their kernels.

Kindly note that the GPU supports float2 (64 bits) and float4 (128 bits) that pack two and four float data, respectively. They support native data-load, but not arithmetic. For half-precision, beyond half2 (32 bits), no other half-based vector data types are supported. This paper explores additional vector types for half-precision.  

\section{Mixed-Precision GNN Training: Analysis, Related Works, and Overview} \label{sec.analysis}
Micikevicius et al.~\cite{micikevicius2017mixed} shows that half-precision is as good as a float for training various models without any accuracy loss. They outline two important points during training: \textit{1)} many forward and backward kernel operations can be carried out in half-precision, hence \textit{state tensors} should also be stored in \emph{half-precision}; \textit{2)} the model weight updates must be in float, which automatically happens through the mixed-precision module of PyTorch. 

Currently, no GNN system has applied these observations successfully. DGL fails to train correctly even on mid-size datasets of Reddit and Ogb-product (Fig.~\ref{obs-accuracy}), while its half-precision SpMM and SDDMM have not shown any performance improvement. Next, we present root cause analysis for abnormal accuracy and system performance issues, while discussing related works.

\subsection{Analysis and Motivation} \label{sec.motivation}
This sub-section analyzes half-precision GNN in more detail and shows that 
\textit{a)} half-precision sparse kernels significantly differ from corresponding dense deep learning kernels. We also show how the current state-of-the-art sparse kernels are not built on the right hardware abstractions. 
\textit{b)} fear of overflow in mixed-precision training forces some PyTorch kernels to be invoked in float data-type due to less understanding about some operations that force additional tensor conversions from float to half and vice versa. 
\textit{c)} half-precision \textit{SpMM} causes value overflow when invoked by mixed-precision training. 
Naively designed GNN systems in such cases would either invoke float-based sparse kernels or perform additional tensor conversion (float to half-precision). 
We also analyze various float-based prior GNN systems to show they would face similar issues if extended to half-precision. We elaborate on these findings next.

\subsubsection{\textbf{Slower Half-Precision Sparse Kernels}}
We postulate that both the \textit{state-of-the-practice} and the \textit{state-of-the-art} for half-precision based GNNs differ from the corresponding dense deep learning model and cannot provide performance benefits at this moment. 

\textit{State-of-the-art} half-precision SDDMM by DGL is hardly better than its corresponding float-based SDDMM, as shown earlier in Fig.~\ref{fig-float-half-sddmmvv}. Specifically, it replaces float with the half-precision data type without any system design change, as they follow Fig.~\ref{fig-half-types}(a). Hence, it does not lead to any performance gain. On the other hand, DGL uses the Cusparse library for SpMM, and Fig.~\ref{fig-float-half-spmm} shows that half-precision SpMM is slower than float-based SpMM. Though it is a closed-source code, profiling reveals to a certain extent that it uses atomic instructions to resolve conflicting writes, which is prevalent in many workload-balanced SpMM (\cref{sec.background.parallel}). Atomic write on half-precision is more costly than the corresponding float data type (\cref{sec.optimization.spmm} and ~\cref{sec.exp.atomic}).

This implies that following a DGL-like strategy of replacing float with half-precision using Fig.~\ref{fig-half-types}(a) (or even Fig.~\ref{fig-half-types}(b)) approach and/or using atomic instruction to resolve conflicting writes to provide half support in other float-based GNN systems~\cite{fu2022tlpgnn,huang2021understanding,wang2021gnnadvisor,wu2021seastar,zhang2022understanding,gnnone,chen2020fusegnn,featgraphsc20,gespmmsc2020, hu2021efficient, liang2020engn, wang2021flexgraph, yan2020hygcn, ye2023sparsetir, zhang2021boostgcn, ferludin2022tf} would unlikely result in any performance gain over their float-based design.




\textit{State-of-the-practice} in half-precision based dense deep learning model focuses on the tensor core, which is well-suited to perform dense matrix multiplication (GeMM). Existing libraries such as CUTLASS by Nvidia are designed to take advantage of tensor cores to perform GeMM. However, the tensor core is unsuited for SpMM due to unstructured sparsity in graph datasets used in GNNs. TC-GNN~\cite{wang2023tc}, which uses tensor cores in GNN, does not perform better than float-based SpMM, as shown by prior works~\cite{Graphpy, GNNBench}.

As this paper presents, good practice would be to use half2 data-type in sparse kernels using cuda cores as opposed to tensor cores, whose native support is available in GPUs (Fig.~\ref{fig-half-types}c).  Earlier, Ho et al.~\cite {ho2017exploiting} have shown that for the dense matrix multiplication before tensor cores became prevalent. 
However, this baseline alone, if developed for sparse kernels, would not be sufficient to achieve the accuracy and envisioned performance as discussed in the rest of the paper. 
Therefore, this paper complements approaches such as CUTLASS libraries as we focus on sparse computations.

\subsubsection{\textbf{Invoking Half-Precision Sparse Kernels with Frequent Data-Conversion}}\label{sec:softmax_exp}

This sub-section shows that \textit{invocation of half-precision sparse kernel in mixed-precision training environment has not been clearly understood}, thereby half-precision sparse kernels may not be invoked or may be invoked with an additional data-conversion as we explain next with a specific example of attention-based GNN, such as GAT, followed by generalizing the observation.

\noindent \textbf{Exponential Operation in GNNs.}
The final edge-level features in attention-based GNN, also called \textit{attention score ($\alpha_{ij}$)}, are derived using an edge-softmax operator. It represents the relative importance of neighbor $j$'s feature to vertex $i$. 
For numerical stability, it is computed using Eq.~\ref{eq-gat}.
Here $e_{ij}$ is the initial feature on each edge, which itself is derived from the features of its source and destination vertices using trainable parameters and an SDDMM variant. It then uses a SpMM variant to get each neighborhood's maximum edge feature ($m_i$).



\vspace{-16pt}
\begin{equation} \label{eq-gat}
m_{i} = {\max_{j \in \mathcal{N}_i} (e_{ij})}, e'_{ij} = \exp(e_{ij} - m_{i}), \alpha_{ij} = \frac{e'_{ij}}{\sum_{j \in \mathcal{N}_i} (e'_{ij})}
\end{equation}
\vspace{-6pt}

However, PyTorch's mixed-precision automatically invokes float-based \textit{exp} in Eq.~\ref{eq-gat} operations instead of half-precision computation.
The exponential operation is immediately followed by an SDDMM variant to implement the division operation of Eq.~\ref{eq-gat}. The output is then sent to SpMMve (not shown in the Eq.) 
Hence, all these follow-up sparse operations receive float tensors due to this upgrade of just one operation (exponential) to float type. 
Hence, GNN systems would either a) perform data conversion from float to half-precision to invoke the half-precision sparse kernels as done in DGL or b) be forced to invoke float-based sparse kernels otherwise. 
The \textit{exp} has already performed a data conversion from half to float on its input.

\textit{Fear of Overflow.} PyTorch mix-precision stipulates that there is a real chance that half-precision is more likely to overflow when computing individual exponential as it can produce values in the range (0, INF). 
However, we point that $e_{ij} - m_i$ can only be a non-positive number in the range of (-INF, 0], hence its exponential cannot cause overflow as the result would only be in (0, 1] which prior works do not take advantage of.

\noindent \textbf{General Trend.}
Exponential is not the only operator where mixed-precision automatically upgrades the incoming half-precision tensor to float before computing. Many operators used widely by GNN systems perform this upgrade~\cite{PyTorchAMP}. The list includes common operations such as cross-entropy, log loss, softmax calculation, summation, etc. So, almost all GNNs should run into these operations where invoking a half-precision sparse kernel would entail a data conversion if the system is not designed accordingly. Even for GCN, CSR-based design of pytorch-geometric~\cite{fey2019fast} shows the data conversion problem where many operations result in half-precision to float data conversion, and subsequent operations are done in float. Its earlier COO-based design often runs out of memory in Reddit, OGB-product, and other mid-size datasets~\cite{Graphpy, chen2020fusegnn, huang2021understanding}, hence, it has not been analyzed.

\subsubsection{\textbf{Value Overflow}}
GCN and GIN use SpMMv 
whose reduction phase causes output to overflow if a vertex has more neighbors, irrespective of the system design in half-precision. \textit{Overflow} condition implies that the value range is too small to hold the output in half-precision, resulting in INF (plus or minus infinity).   
Even modest-sized datasets, such as Ogb-Product and Reddit, have many vertices with large enough neighbors. 
The INF then produces NaN in the follow-up operations, e.g., the softmax layer, which operates on two INF. Hence, approaches of Fig.~\ref{fig-half-types}(b) and (c) would always result in value overflow in SpMM when neighbors are large. However, if one chooses to use Fig.~\ref{fig-half-types}(a), SpMM would still overflow during reduction as it would likely produce a float value that overflows when converted to half-precision.


Next, we show that GNN models offer built-in mechanisms to guard against overflow but have not been used by prior works.


\noindent \textbf{GCN.}
Eq.~\ref{eq-gcn} shows its simplified computation.
where, \( {\mathbf{D}} \) is the degree matrix of graph \({\mathbf{A}} \). \( \mathbf{H}^{(l)} \) is the activation of the previous layer, and \( \mathbf{W}^{(l)}\) is the weight matrix of linear layer. The linear layer transforms the activation of the previous layer, which then serves as a dense input matrix to the SpMMv.
This also includes a \textit{degree-norm} scaling. GCN supports 3 different kinds of degree-norm: 1) \textit{left}: divide the input dense tensor of SpMMv with degree; 2) \textit{right}: divide the output tensor of SpMMv with degree; and 3) \textit{both}: divide the input and output dense tensors of SpMMv with the square root of degrees separately, which is shown in Eq.~\ref{eq-gcn}.


\vspace{-11pt}
\begin{equation} \label{eq-gcn}
\mathbf{H}^{(l+1)} = \sigma \left({\mathbf{D}}^{-\frac{1}{2}} {\mathbf{A}} {\mathbf{D}}^{-\frac{1}{2}} (\mathbf{H}^{(l)} \mathbf{W}^{(l)}) \right)
\end{equation}
\vspace{-6pt}

The degree-norm scaling adds stability to feature values. To illustrate, consider the frequently used degree-norm, right, which divides the output vertex features of SpMM by the vertex degree. 
Hence, the degree-norm can bring the SpMMv output within the half-precision range. However, the current systems perform degree-norm after the reduction is \textit{over}, by then overflow has already happened.
Some prior float-based GNNs~\cite{wang2021gnnadvisor} have fused the degree-norm with {SpMMv}. However, our analysis shows that such approaches are not helpful because they also do degree-norm after the reduction phase. Hence, prior works cannot protect against overflow in half-precision computation.

Kindly note that if one uses the left degree-norm, there will be no overflow. However, during backward computation, the degree-norm happens after SpMMv (right), where it is likely to overflow.

\noindent 
\textbf{GIN.}
GIN has many variations, we focus on two variations. a) In the most common version, there is no stabilization operation like the degree-norm discussed for GCN. 
Therefore, there is no method to protect against the value overflow if one uses half-precision. 
b) DGL offers a few additional reduction operations in addition to the `\textit{sum}' for the SpMM operation in GIN. One of those options is `\textit{mean}' reduction, which is the same as SpMM followed by degree-norm as discussed in GCN. 
However, this DGL option is implemented the same way as GCN, i.e., the degree-norm is called after \emph{SpMM} for forward computation. Consequently, this version of GIN is susceptible to the same overflow issue as GCN, while GIN implemented by other float-based GNN systems would always overflow.

\subsection{{\halfgnn} Overview} \label{sec.overview}
The proposed {\halfgnn} tackles the performance and accuracy issues together.
First, {\halfgnn} adopts the half2-based design to arrive at a better half-precision baseline and shows unique challenges in its adoption. It proposes \textit{edge-feature mirroring}, \textit{feature padding} for odd vertex features, and introducing \textit{sub-warps} to impart better thread utilization to adopt half2. 
However, reduction-related design, including reliance on atomic write, continues to cause performance bottlenecks.
Lastly, the resultant design would still suffer from the overflow issue, whose root cause is reduction-related designs. 

To this end, {\halfgnn} further proposes \textit{half4} (64 bits) and \textit{half8} (128 bits) vector data types, packing four and eight half-precision data respectively. These two vector types natively support data loading. Utilizing them improves SDDMM performance by minimizing the inter-thread communication cost, which indirectly improves the instruction-level parallelism to offer better data-load performance. 
For SpMM, \textit{discretized reduction} scaling is proposed where only a finite batch of neighbors is reduced and degree scaled before moving to the next batch of neighbors. The design allows us to implement multiple workload-balancing solutions.
We then introduce a staging buffer to remove atomic write, the root cause of the performance bottleneck in half-precision SpMM. This design also protects against overflow in SpMM due to the discrete nature of reduction and degree-norm scaling.

The proposed solutions are generic and apply to both the edge-parallel and the vertex-parallel sparse kernels. E.g., the discretized reduction is very well suited to provide workload balancing in edge-centric and vertex-centric sparse kernels. 
However, {\halfgnn} recommends an edge-parallel solution for the best performance, which we discuss in detail, while we also show its applicability in a vertex-parallel solution(\cref{sec.exp.microbenchmark}) of Huang et al.~\cite{huang2021understanding}.

\section{Adapting Half2 Data-Type} \label{sec.half2}
This section focuses on adapting half2 to form a baseline, which we can analyze to improve performance and accuracy.
We discuss the challenges and techniques in adopting the half2 data type. We strive to incorporate state-of-the-art techniques from float-based sparse kernel design, such as workload balancing, while developing the baseline. However, the focus is also to show the applicability of proposed techniques to edge-parallel and vertex-parallel design paradigms to claim general applicability.

\subsection{Better Data-Load using Half2 Data Types} \label{sec.load}
Substituting half in place of float would always compromise the data load performance.
In such an approach, each thread brings one feature (2-byte) from GPU global memory to register for computation, hence a warp performs only 64 bytes of data-load. In comparison, a float-based system issues 128 bytes of data load per warp, which achieves full coalesced memory access. 
This is evident from the DGL SDDMM result (Fig.~\ref{fig-float-half-sddmmvv}) that does not show any runtime improvement as it merely replaces the float with the half data type.

To fully utilize the memory transfer capability, {\halfgnn} adapts the half2 data type, provided natively by CUDA, which prior GNN systems have not used. We propose a split data-load phase to overcome the challenges associated with adopting \emph{half2}. 


\subsubsection{\textbf{Phase1: Explicitly Loading NZE and Edge-Features.}} \label{sec.load.pahse1}
We observe that each NZE and its edge-feature are used $|F|$ times (feature length of the vertices) during computation. 
To illustrate, the role of NZE is to know the column (vertex ID) so that its vertex features (vector) could be fetched for SpMM and SDDMM. As different warp threads fetch different features of the same column, the column ID could be used by them if we fetch it once and cache it.
Similarly, the edge-feature of the NZE performs a dot product with all the corresponding vertex features in SpMM. 

However, in a naive feature-parallel method, each thread of the warp would load the same NZE and edge-feature, leading to repeated and non-coalesced access.
{\halfgnn} deploys an \textit{explicit edge-parallel method} to load and cache both of them to the GPU shared-memory. This allows their data-load to be performed in a coalesced fashion, while caching in shared memory achieves the effect of broadcasting, allowing feature-parallel threads to read locally when needed in the next phase.  

Further, {\halfgnn} treats two contiguous \emph{half-precision} edge features as one \emph{half2} feature. A simple type-casting is all that is needed to cast the half features to the half2 data type, if designed carefully (see feature padding for challenges involved). Thus, a warp can issue 32 \emph{half2} values, leading to 128 bytes of data-load. 
To take advantage of this phase, at least 64 edges must be allocated to each warp to process. This would be equivalent to 32 half2 (128 bytes) features.

\subsubsection{\textbf{Phase2: Implicit Vertex Feature Loading.}} \label{sec.load.phase2}
To load vertex features, threads need to know the column ID of each NZE, which is already present in the shared memory thanks to phase 1. Vertex features are stored in row-major order, and accessing 2 of them in a group is straightforward. As long as the feature length is a multiple of 2, a simple type-casting of the features tensor to half2 allows us to use the half2 data type for data-loading. 
This allows 1 warp thread to carry out data load for 2 features at the same time, instead of 1. So, considering 32 threads in each warp, it can handle 64 features. So, it doubles the number of features that can be loaded in parallel within each warp over a float-based system. 

\noindent \textbf{Feature Padding for Odd Feature Lengths.}
Kindly note that the feature length is a configurable parameter in GNN, feature lengths such as 32, 64, etc., are common configurations. However, such lengths could be odd in the last layer of GNNs when the classification category itself is odd. E.g., Reddit has 41 classes to predict. 
In such cases, a simple type-casting half to half2 does not work due to memory alignment issues, where hardware would not allow accessing half2 values whose address is not a multiple of 4 bytes. In such a case, we simply use the next even number as the class count without any issues. 

\noindent \textbf{Sub-Warps.}
We observe that vertex-level features are loaded in a feature-parallel way within a warp. Hence, the threads being used are now half of what would be used in float-based sparse kernels. 
E.g., when the feature-length is 32, we only need 16 threads of any warp to when using \textit{half2},. 
Hence, we process two non-zero columns at one time by the warp by dynamically creating two sub-warps within a warp. This allows us to use all the threads of the warp, leading to better thread utilization. 
Indeed, we generalize this based on the feature-length, and hence the number of sub-warps within a warp depends on the feature length. E.g., in the case of 16 as feature length, we create 4 sub-warps.

\subsection{Faster Computation using Half2 Data-Type} \label{sec.load.compute}

Fig.~\ref{fig-half-types}(a) approach illustrates that native arithmetic using operators such as `+' or `*' causes implicit data conversion of half-precision operands to float, and the operation is performed in float. Similarly, CUDA intrinsic APIs for half computation avoid this data-type conversion but achieve the same throughput as the float (Fig.~\ref{fig-half-types}(b)). Our approach is to follow Fig.~\ref{fig-half-types}(c) as CUDA intrinsic natively supports \emph{half2} arithmetic, while our data is already in half2. 
This approach doubles the arithmetic computation throughput but faces a unique challenge, discussed next.

\noindent
\textbf{Edge-Feature Mirroring.} The dot-product phase in SpMM multiplies the edge feature (in half) by the vertex features (in half) of the non-zero column of the edge. However, usage of half2 packs two edge features as one half2. 
This means that only one of the halfs of the edge feature should do the dot product with both halfs of the half2 vertex features for any given edge. 
To illustrate, Fig.~\ref{fig-background} (d) for SpMM shows that only $W_{21}$ should be multiplied with \{$X_{10}$, $X_{11}$\}, but edge-feature data loading brings $W_{21}$ together with $W_{23}$ in our edge-centric design as half2 data-type. So, performing dot product naively would also multiply $W_{23}$ with \{$X_{10}$, $X_{11}$\}, which is wrong.

Hence, in phase-1 of data-load (\cref{sec.load.compute}), after data loading edge features and before caching them in shared memory, we extract each half from half2 edge features, and is mirrored to create new half2 edge features which are then directly stored in the shared memory for caching. To illustrate, the first half2 word contains \{$W_{21}$, $W_{21}$\} (to be multiplied with \{$X_{10}$, $X_{11}$\}), and the next half2 contains \{$W_{23}$, $W_{23}$\} (to be multiplied with \{$X_{30}$, $X_{31}$\}).

Performing this mirroring does not introduce any additional data-load as it happens after the data-load of edge features. Hence, it ensures that edge features can still use the half2 data type for the data-load and dot product phases. The mirroring allows us to utilize the approach of Fig.~\ref{fig-half-types}
(c) resulting in doubling the arithmetic throughput to perform the dot product compared to float-based computation.

\begin{figure*}[t]
  \centering
  \includegraphics[scale= 0.32]{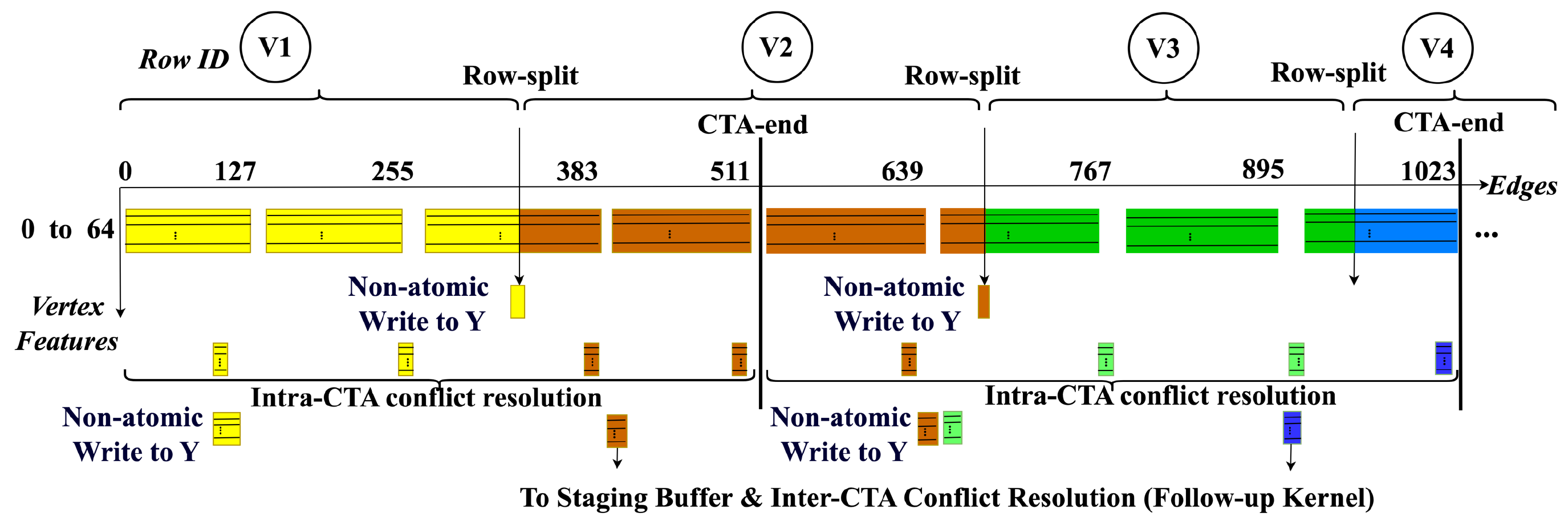} 
  \vspace{-6pt}
 \caption{\small {Reduction in Edge-Parallel SpMM of {\halfgnn} showing 2 CTAs, each containing 4 warps. Each warp handles 128 edges. Vertex feature length is 64 ($|F|$). \textit{a)} The dot products are locally aggregated by each warp by performing a running reduction. \textit{b)} The non-conflicting reduced output is written directly in the output tensor ($Y$). \textit{c)} The conflicting reduced output ($|F|$ per warp) is first aggregated within CTA using shared memory, which generates many non-conflicting writes and one final conflicting write ($|F|$ per CTA). The latter is sent to the staging buffer, which is handled by a follow-up kernel to write them to the output tensor ($Y$).} See \cref{sec.spmm.atomic} for details on the method.}
  \label{fig-arch}
 \vspace{-6pt}
\end{figure*}

\section{Reduction Centric Design} \label{sec.optimization}
The developed baseline using the half2 data type still does not achieve the envisioned performance and accuracy goals without focusing on a design centered around the reduction phase. 
This section discusses the SDDMM and SpMM design to overcome the issue, which changes the data load and computation phases to accommodate the concerns arising from their reduction design.
  
\subsection{SDDMM: More Vector Data-Types} \label{sec.optimization.sddmm}

SDDMM reduction happens on the dot products
along the feature dimension. As feature parallelism is used to load features and compute dot products, the reduction requires inter-thread communication to generate a half feature per NZE (edge).

\subsubsection{\textbf{Observation}}
Inter-thread communication implies a memory barrier that does not allow the compiler to reorder memory instructions (e.g., data loads) across the barrier. The barrier makes warp threads wait until all threads finish their work. Hence, threads only issue one data load of size 32 bits (half2) before synchronizing at the barrier. 
Further, a warp performs four rounds of inter-thread communication to perform reduction on 32 half2 values. 

Thus, the baseline version of SDDMM using \textit{half2} data type does not have an optimal design as it is not sufficient to realize the full data-load potential of GPUs, as we show in the evaluation (\cref{sec.exp.microbenchmark}).
{\halfgnn} proposes two new vector data types to fully optimize the performance, as discussed next.

\subsubsection{\textbf{Half4 and Half8}}
We propose, for the first time, two new vector data types of \emph{half4} and \emph{half8} that pack four and eight half-precision data.
To provide their native support, \emph{half4} internally uses \emph{float2} data-type to issue data-load, i.e., each warp can issue 256 bytes of data-load in a single data-load instruction where each warp thread loads 4 half-precision data. Similarly, \emph{half8} internally relies on \emph{float4} data type to issue 512 bytes of data-load using a single data-load instruction. Float2 and float4 are natively supported by GPUs for data-load purposes and not for computation. Thus, half4 and half8 use half2 for arithmetic.

To this end, {\halfgnn} offers a range of data types to researchers where continuous memory locations of features can be accessed, using a pointer data type of  \textit{half, half2, half4, and half8} for read-write operation, provided memory constraints are maintained. For example, if we have a pointer pointing to an array of data-type half, we can easily switch its pointer type from \emph{half} to \emph{half2, half4}, and \emph{half8} if the array size is a multiple of 2, 4, and 8 respectively.


\subsubsection{\textbf{Half8 for SDDMM}}
{\halfgnn} SDDMM follows an edge-parallel design and continues to have the two-phase data-load design.
However, it uses \textit{half8}  in phase 2 of data-load. Using \textit{half8}, each warp thread loads 8 half features of the non-zero columns in phase 2 of the data-load (\cref{sec.load.phase2}).  It then performs a dot product of the vertex features of rows and column IDs of each NZE. The dot product phase uses half2 because arithmetic on half8 is not supported by hardware.
The usage of \emph{half8} also minimizes the inter-thread communication. To illustrate, consider 32 as feature-length, where only 4 threads are needed. This requires only two rounds of inter-thread communication. The half2-only solution requires four rounds of inter-thread communication, while a half-based SDDMM would use five rounds of inter-thread communication.

The major reason for the increased performance of half8-based SDDMM is the lesser impact of the memory barrier on the data load. In other words, we now effectively issue a load of 4 \emph{half2} features before observing the memory barrier compared to issuing just one half2 feature load in half2-only SDDMM.
The sub-warp concept (\cref{sec.load}) helps us utilize the idle threads when using these vector types.




Please note that, like the float2 and float4, \emph{half4} and \emph{half8} do not improve computation throughput compared to \emph{half2}. Hence, even though we provide arithmetic on half4 and half8 data types, it internally relies on half2 arithmetic. We continue to rely on feature padding to use the newly proposed data type when the feature lengths are not appropriate, such as in the last layer of GNN, where the feature length is determined by the number of classes to predict.



\subsection{SpMM: Discretized Reduction Scaling} \label{sec.optimization.spmm}

The reduction in SpMM is performed in the neighborhood dimension, so we need inter-thread communication only when the whole row is assigned to more than one warp-- a norm in most workload-balanced SpMM designs~\cite{huang2021understanding,wang2021gnnadvisor,gnnone,yang2018design,wang2023tc}. 
However, inter-thread communication is replaced with atomic write in the prior float-base workload-balanced SpMM solutions, and is a major performance bottleneck. This sub-section shows a close association between this approach and the value overflow issue (\cref{sec.motivation}), to solve both together.


\subsubsection{\textbf{Observation}}

Fig.~\ref{fig-arch} shows the edge-parallel {\halfgnn} SpMM incorporating the proposed discretized reduction scaling, which we explain shortly. The proposal is based on our observation that almost all state-of-the-art workload-balanced SpMM follow these rules: 



\begin{enumerate}[nosep]
    \item \emph{Every edge is uniquely mapped to a single warp and consequently to a CTA, where its threads load vertex features in a feature-parallel way.}

    \item \emph{More than one edges are allocated to the same warp or CTA, whereby threads handle them in a loop. The consecutive edges follow the spatial ordering, hence their row ID may be the same or in monotonically increasing order if a row split is observed. 
    }
    \item {\emph{Edges distributed to consecutive CTA may share the same row ID if a row has large non-zero columns.}}
\end{enumerate}



Rule 2 implies that a warp or CTA handles only a group of edges at any time. We will explain soon how this is done for different computation paradigms of edge-parallel and vertex-parallel SpMM.
But first, this count is small enough that we observe that SpMM forms a batching of edges or neighbors. Hence, \textit{we claim the computation is discretized}, which we exploit to remove atomic writes and protect value overflow. 

The key difference between edge-parallel and vertex-parallel is that row-split is observed in the former within a warp, as shown for the third warp of the first CTA in our edge-parallel approach of Fig.~\ref{fig-arch}, as it allocates every warp an equal number of edges to compute irrespective of their source vertex ID (row ID). 
On the other hand, workload-balanced vertex-parallel SpMM does not force row-split. For example, Huang et al.~\cite {huang2021understanding} and GNNAdvisor~\cite{wang2021gnnadvisor} allocate only a group of 32 neighbors to each warp. The last warp may handle fewer than 32 neighbors, as the vertex degree may not be a multiple of 32.
Even vanilla vertex-parallel SpMM that offers no workload balancing, such as GE-SpMM~\cite{gespmmsc2020}, has an implicit grouping where each warp handles 32 neighbors of any vertex before processing the next 32 neighbors for the same vertex in the next iteration. The last iteration may process fewer than 32 neighbors, depending on the vertex degree.
We, of course, proposed to handle at least 64 edges in place of 32 in such designs (\cref{sec.load.pahse1}), nonetheless, they all follow a discretization of computation in place, which we exploit next to protect against value overflow.


\subsubsection{\textbf{Protecting Against Value Overflow}} \label{sec.discrete}

Our proposed method is based on the observation of how the reduction is performed currently and how it can be tailored to solve the problem. In effect, SpMM + degree-norm is treated as a custom SpMM kernel with `mean' as a reduction operation, thus solving the problem for both GCN and GIN  models.

We term the currently used reduction approaches as \textit{post-reduction} scaling (degree-norm scaling applied once at the end of the reduction of the full neighborhood of vertices), performing exactly one division per vertex at the end of the reduction. 
The other end of the spectrum would be to perform degree norm scaling before reduction, i.e.,  divide each dot product by degree and then perform reduction, called \textit{pre-reduction} scaling. This would perform exactly $|N(v)|$ count of division for each vertex, avoiding INF generation, but introducing more arithmetic operations. While a few values might vanish (underflow) due to division at each dot product. 

The proposed \textit{discretized reduction} scaling shares the pros of both and the cons of none.
This applies degree-norm scaling after reduction on a batch of neighbors for each vertex (observation rule 2).  
To illustrate, we perform the degree-norm scaling at the end of the reduction for each batch. Thus, discretizing the reduction never risks getting INF as long as the maximum number of neighbors in the batch is small, which is the case as discussed in the observation. 
Thus, the proposed design changes are minimal and effective. 

\vspace{-12pt}
\begin{equation} \label{eq-gin}
h_u = \phi\left((1 + \epsilon) \cdot x_u + \sum_{v \in N_v} x_v\right)
\end{equation}

\noindent \textbf{Additional Overflow in GIN.}
\emph{Discretized Scaled SpMM} completely removes accuracy issues for GCN, but it is still not enough for GIN.
In GIN aggregation, we observe 2 parts, one is the right representing unweighted \emph{SpMM} being added to scaled source node features as shown in Eq.~\ref{eq-gin}.
The overflow issue does occur when adding self-features for sources to the SpMM output. We introduce a non-learnable parameter $\lambda$, to scale the aggregation down. Ideally, $\lambda = 0.1$ worked fine for all our robust testing. So, we propose the following kind of aggregation for GIN in \emph{half-precision}:

\vspace{-12pt}
\begin{equation}
h_u = \phi\left((1 + \epsilon) \cdot x_u + \lambda \cdot \left( \frac{1}{|N_u|} \sum_{v \in N_u} x_v\right)\right)
\end{equation}



\subsubsection{\textbf{Non-Atomic Writes}}  \label{sec.spmm.atomic}



The discretized reduction also introduces conflicting writes for which prior float-based works have relied on atomic writes for the data-store phase, which is very costly for half2-based atomic writes as per our measurements (\ref{sec.exp.atomic}). 

\noindent \textbf{Conflict Write Identification and Handling.}
We now lay out when the conflicting writes occur and how to resolve them in our edge-parallel SpMM (Fig.~\ref{fig-arch}). 

\begin{enumerate} [nosep]
\item if one or more rows are contained exclusively within a warp, there is never any conflicting write, and hence no atomic write is needed. 

\item If one or more rows are contained exclusively within a CTA, then an {intra-CTA communication} among its warps can resolve those conflicting writes locally within the CTA to generate non-conflicting writes. We will discuss this soon.

\item Only when a row extends to more than one CTA, the conflicting writes cannot be resolved locally. \textit{a)} Kindly note that the reduced values generated in the case of row-split within a warp are non-conflicting if we never perform any other conflicting write to the output tensor $Y$ directly. That is, the third warp in Fig.~\ref{fig-arch} is non-conflicting for row ID $V1$ if the first and the second warps do not perform their write (for $V1$) to $Y$ directly. \textit{b)} The reduced values at \textit{warp-end} (non-row-split case) are the ones that are always conflicting.

\end{enumerate}

Fig.~\ref{fig-arch} shows the details: \textit{a)} the thread-local reduced value is written non-atomically to the output tensor directly in case of \textit{row-split} (only) as shown for the yellow row (Case 3(a) above). \textit{b)} The reduced values at \textit{warp-end} (non-row-split case), i.e., case 2 and case 3(b) above, are the ones that are resolved first locally within a CTA by using shared memory. This resolves all conflicting writes within CTA, and finally generates one conflicting write of size $|F|$ per CTA: the ones that can only be resolved by looking into the writes by the next CTAs. The former is written directly to $Y$ while the latter is written to the staging buffer.
The staging buffer is then re-read in a separate follow-up kernel to write to the output tensor.

\begin{figure*}[t]
  \centering
  \includegraphics[scale=0.59] {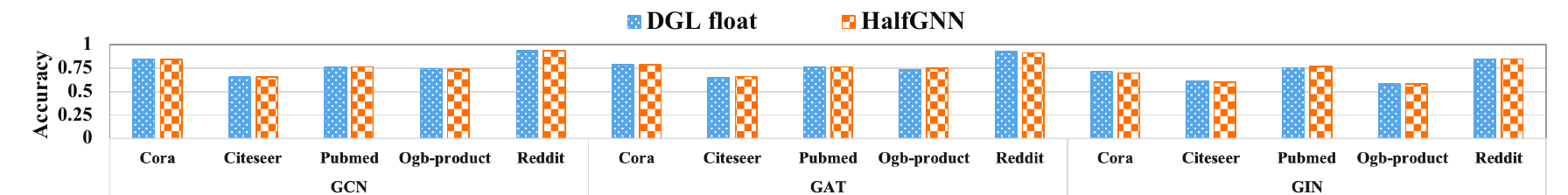} 
  \vspace{-12pt}
  \caption{\small {\halfgnn} achieves the same accuracy as float-based DGL, confirming the applicability of the proposed techniques. Only labeled datasets (Table~\ref{table-dataset}) are used to show accuracy.} 
  \label{exp-dgl-halfGNN-acc}
  \vspace{-6pt}
\end{figure*}

Kindly note that current libraries, such as Nvidia's CUB, offer intra-CTA communication only for scalar dot products for reduction, and are insufficient for exchanging vector dot products, as needed for our purpose. 
We propose a new design for intra-CTA communication, which also handles the case of sub-warp ({\cref{sec.load}}). In such a case, not all threads need to communicate with every other thread; the communication happens between a thread of a sub-warp to the same thread ID of all other sub-warps of the CTA, so that the same feature gets reduced. Hence, if there are 8 sub-warps within a CTA, we only perform three rounds of communications within the CTA. 
The proposed library relies on the cached row ID stored in the shared memory to know if the dot products stored in a sub-warp should be merged with the dot products of another sub-warp.

\noindent
\textbf{Staging Buffer Size.} This buffer stores the \emph{conflicting write} of each CTA. Hence, its size is $|CTA|\times|F|$ features. 
This buffer requires little memory and is reused by each SpMM. Further, we already save a lot of memory by using \emph{half-precision} for state tensors, hence this small buffer allocation is not a major concern. 

\subsection{Shadow APIs}

In mixed-precision training, DL frameworks invoke float versions of many operations~\cite{PyTorchAMP}, e.g., exponential (\cref{sec:softmax_exp}). However, if we are sure that such operations would produce the output within the numerical range of half-precision, we advocate using the half-precision version. This guarantees an optimal number of kernel calls are \textit{half-precision} and the least number of float conversions for maximum performance.  

However, DL frameworks, such as PyTorch, do not automatically invoke such an operation in half-precision~\cite{PyTorchAMP}. To this end, we provide similar APIs, called \textit{shadow API}, that guarantee that the half-precision version would be invoked if the input tensor is of half data type, and the float version would be invoked if the input tensor is of float type. Hence, one can always invoke the proposed shadow API inside the graph convolution layer if they are sure that overflow would not incur in half-precision.

\subsection{\textbf{Generality of the Solution }} \label{sec.general}
Most of the proposed ideas are equally applicable to edge-parallel, vertex-parallel, and their variants, though {\halfgnn} is an edge-parallel solution. For example, usage of vector data-types (half2, half4, and half8), edge feature mirroring, feature padding, sub-warps, discretized SpMM, removal of atomic writes, and proposed shadow APIs apply to all.  
The key differences are in phase 1 of data load, where edge features are loaded, and in the handling of conflicting writes, which are discussed next.

\noindent \textbf{Data-Load of Edge features.}
An edge-parallel system allocates an equal number of edges to every warp. However, vertex-parallel systems allocate unequal edges per warp as they never experience row-split. For example, a few workload-balanced vertex-parallel SpMM~\cite{huang2021understanding,wang2021gnnadvisor} allocates a maximum of 32 neighbors of a vertex to each warp to compute on. However, the last warp allocated to process such vertices may process fewer than 32 neighbors, depending on the degree of the vertex. Similarly, warps that handle vertices with a degree smaller than 32 process fewer neighbors. Though their allocation strategy impacts their workload balancing, it affects the data load of edge features, which 
does not always start at an even offset. 
To illustrate, if the first vertex has 11 neighbors, the memory pointer pointing to the edge features corresponding to the next vertex's neighborhood cannot be type-cast to half2 due to the memory alignment issue, leading to complexity in using half2. 



\noindent \textbf{Conflict Write Handling.}
Our non-atomic design principle involves intra-CTA and inter-CTA conflict resolution, with a 2-stage kernel design. This applies to edge-parallel design as {\halfgnn} itself is edge-parallel. This applies to vertex-parallel design as well, except that we do not require intra-warp conflict resolution because a row split is not observed in this case. However, intra-CTA and inter-CTA remain in place.

\noindent \textbf{Implementing Half-precision SpMM of Haung et al.~\cite{huang2021understanding}.}
To tackle the memory alignment issue, we start edge-feature fetch from one position earlier if needed, and handle this carefully during edge-feature mirroring to write the correct value in the shared memory. 
We apply the rest of the vectorization design and non-atomic write. \cref{sec.exp.huang} shows the benefit of this design.

\begin{table}[t]
\small
\caption{ \small Graph datasets. * denotes \textit{labeled} dataset while others have generated features and labels. $|F|$ is the Input feature-length, $|C|$ is the Prediction categories (or output feature-length), while the intermediate feature-length is set as 64.} 
\centering 
\begin{tabular}{l r r r r} 
    \hline\hline 
    Graph Dataset & $|V|$ & $|E|$& $|F|$ &  $|C|$ \\
    \hline 
    Cora (G1)* & 2,708 &  10,858 & 1,433 & 7\\ 
    Citeseer (G2)* & 3,327 & 9,104 & 3,703 & 6\\
    PubMed (G3)* & 19,717 &  88,648 & 500 & 3\\
    Amazon (G4)  & 400,727 &  6,400,880 & 150 & 7\\ 
    Wiki-Talk (G5) & 2,394,385	& 10,042,820 & 150 & 7\\
    RoadNet-CA (G6) & 1,971,279&	11,066,420 & 150 & 7\\
    Web-BerkStand (G7) &685,230	& 15,201,173  & 150 &  7\\ 
    As-Skitter (G8)   &1,696,415 &	22,190,596 & 150 & 7\\    
    Cit-Patent (G9) & 3,774,768 & 33,037,894 & 150  & 7\\
    Sx-stackoverflow (G10) &2,601,977 &	95,806,532 & 150 & 7\\
   
    Kron-21 (G11) &2,097,152 & 67,108,864 & 150 & 7\\
    Hollywood09 (G12) &1,069,127 & 112,613,308 & 150 & 7\\
    Ogb-product (G13)* &2,449,029 & 123,718,280 & 100 & 47\\
    LiveJournal (G14) &4,847,571 &	137,987,546 & 150 & 7\\
    Reddit (G15)* &232,965 & 114,848,857 & 602 & 41\\
    Orkut (G16)  &3,072,627 & 234,370,166 & 150 & 7\\
    
    \hline 
\end{tabular}
\label{table-dataset} 
\end{table}

\section{Evaluations} \label{sec.exp}

We comprehensively evaluate the proposed optimizations and perform a comparative study over prior GNN works. All the experiments are conducted on an NVIDIA A100 GPU (40GB of GPU memory) using 16 datasets mentioned in Table~\ref{table-dataset}. Only 5 datasets (Cora, Pubmed, Citeseer, Ogb-product, and Reddit) are labeled.
{\halfgnn}\footnote{Code can be accessed from \url{https://github.com/the-data-lab}} is integrated into the GNNBench benchmarking platform~\cite{GNNBench} to avoid pitfalls that have been discovered recently~\cite{Graphpy}. The platform allows to use of non-labeled datasets by using generated labels and features, whose dimensions are listed in Table~\ref{table-dataset}, and are used for only performance measurements due to the limited number of labeled datasets. It also recommends that runtime results on smaller datasets (G0--G2) are not reliable, hence, they are used only for verifying accuracy.

We focus on GCN, GAT, and GIN models.
These models contain intermediate feature lengths of 64. The intermediate feature length is important as the sparse kernels usually have this feature length during computation. The original feature length is usually projected to this intermediate feature length through GeMM (a part of the linear layer). Also, the SpMM and SDDMM of the last layer usually use a feature length equal to the prediction categories ($|C|$ in Table~\ref{table-dataset}).  Thus, our evaluations include various feature lengths for sparse kernels.
All GNNs were trained for 400 epochs.

Kindly note that DGL is the state-of-the-art GNN system for half-precision, though it suffers from value overflow in GCN and GIN models, which directly uses Cusparse half-precision SpMM. 
Pytorch-Geometric~\cite{fey2019fast} frequently runs out of memory when using its COO-based design. Its recent implementation that uses CSR format supports GCN-like models only, which also suffers from data conversion issues, while GAT continues to rely on the COO format~\cite{Graphpy}. Hence, we mostly use DGL as the baseline and rely on various micro-benchmarks and performance counters to justify the proposed optimizations. 
We also used Huang et al.~\cite{huang2021understanding} SpMM, a workload-balanced solution, which is the state-of-the-art vertex-parallel float-based SpMM, to show the applicability of the proposed techniques to the vertex-parallel computation paradigm.



\subsection{GNN Training}
This sub-section evaluates training results, showing accuracy, runtime, and memory usage. 

\subsubsection{\textbf{Training Accuracy}} \label{sec.exp.accuracy}

    

As we highlighted earlier in Fig.~\ref{obs-accuracy}, the accuracy of DGL for half-precision is much lower than the float-based models.
Hence, we use DGL single-precision as the baseline in Fig.~\ref{exp-dgl-halfGNN-acc} to show that {\halfgnn} achieves similar accuracy to the baseline. 
The improvement in accuracy is less than 0.3\% for all except Pubmed in GIN for which it is less than 1.0\% improvement. 
Firstly, half-precision works as a regularizer, resulting in minor accuracy improvements as discussed by mix-precision training~\cite{micikevicius2017mixed}.
Secondly, we have made a minor adjustment to protect the additional overflow in GIN as discussed in \cref{sec.discrete}. 
\textit{These accuracy results imply that half-precision is a viable option in GNN in place of the float} if our proposed techniques are utilized. 

\noindent \textbf{Overflow Protection {is} the Key.}
Overflow protection through \emph{discretized reduction} scaling in \emph{SpMM} by {\halfgnn} is the key reason to achieve accuracy similar to DGL-float. 
To confirm this result, we replaced the discretized reduction with the usual reduction so that overflow would not be protected. This system did result in DGL-half like abnormal accuracy in GCN and GIN.

\begin{figure*}[t]
  \centering
  \includegraphics[scale=0.53]{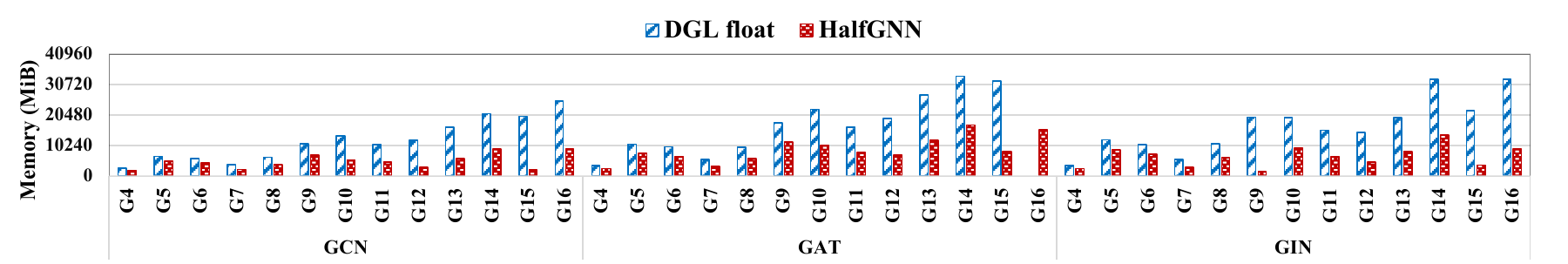} 
 \vspace{-6pt}
  \caption{\small {\halfgnn} requires much less memory to train compared to DGL float. (Lower is better)}
  \label{exp-dgl-halfgnn-memory}
  \vspace{-6pt}
\end{figure*}

\subsubsection{\textbf{Memory Consumption}}
{\halfgnn} also achieves 2.67$\times$ less memory consumption on average than float-based DGL, considering all 3 GNNs as plotted in Fig.~\ref{exp-dgl-halfgnn-memory}. While it is clear that half-precision would improve memory usage, one must note that the saving is not purely due to moving to half-precision, as DGL consumes excessive memory due to its design choices and has framework-memory overhead as shown recently~\cite{GNNBench}.

\subsubsection{\textbf{Training Runtime}} \label{sec.exp.runtime}

\begin{figure}[t]
  \centering
  \includegraphics[width=0.49\textwidth,]
   {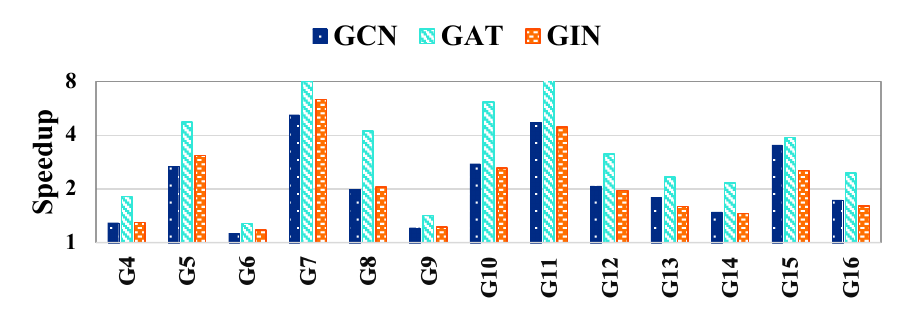}
 \vspace{-11pt}
  \caption{\small{{\halfgnn} training runtime speedup over DGL-half for feature size 64. (log scale, higher is better)}}
  \label{exp-halfgnn-speedup-half}
  \vspace{-11pt}
\end{figure}

\begin{figure}[t]
  \centering
  \includegraphics[width=0.48\textwidth,]
  {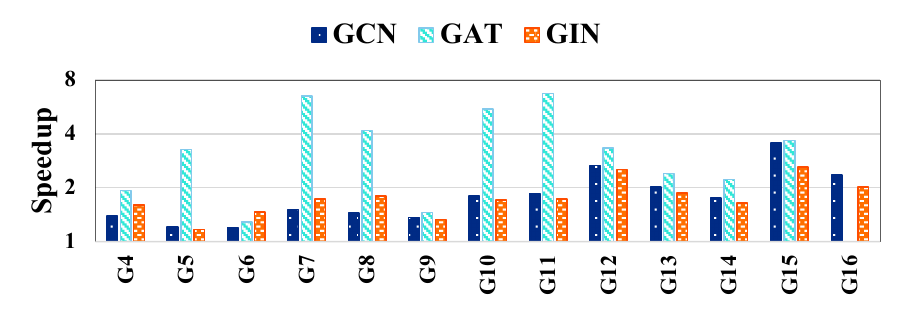}
  \vspace{-11pt}
  \caption{\small{{\halfgnn} training runtime speedup over DGL-float for feature size 64. DGL runs out of memory for GAT G16. (log scale, higher is better)}}
  \label{exp-halfgnn-speedup}
\end{figure}

Fig.~\ref{exp-halfgnn-speedup-half} shows that {\halfgnn} achieves 2.44$\times$, 3.84$\times$, and 2.42$\times$ speedup for GCN, GAT, and GIN, respectively, compared to DGL-half.
Compared to DGL-float, {\halfgnn} achieves $1.85\times, 3.55\times$, and $1.78\times$ average speedup for GCN, GAT, and GIN, respectively, as shown in Fig.~\ref{exp-halfgnn-speedup}. 


{\halfgnn} speedup is greater compared to DGL-half than DGL-float. This confirms our earlier observation where Fig.~\ref{fig-float-half-spmm} and Fig.~\ref{fig-float-half-sddmmvv} have shown that half-precision based DGL kernels are either slower (for SpMM) or have similar performance (for SDDMM) compared to the float-based DGL. Hence, by showing better performance over DGL-Float (faster and correctly executed baseline), {\halfgnn} truly advances the state-of-the-art.

The reason for the faster training time is due to the better sparse kernels of {\halfgnn}, specifically due to half-precision specific performance optimization to the sparse kernels as shown in \cref{sec.exp.kernel}. Kindly note that GNN training involves many other kernels such as linear layer, activation, bias, softmax, etc., where baseline and {\halfgnn} use the same implementation of PyTorch. Hence, a sparse kernel speed-up (\cref{sec.exp.kernel}), which is large for half-precision kernels of SpMM and SDDMM, still results in a significant average training time speed-up here.   

\subsection{Kernel Evaluation} \label{sec.exp.kernel}

This section focuses on individual kernel performances compared to established DGL/Cusparse kernels.

\begin{figure}[b]
  \centering
  \includegraphics[width=0.49\textwidth,]{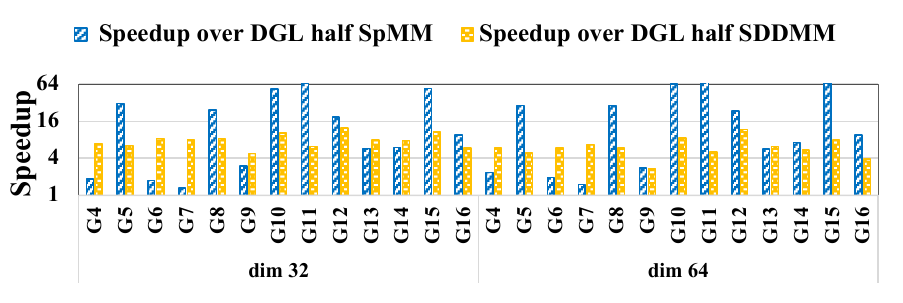} 
  \vspace{-24pt} 
  \caption{\small{{\halfgnn} achieves $22.89\times$ average speedup over DGL half-precision SpMM (Cusparse), and $7.12\times$ average speedup over DGL half-precision SDDMM. Speedup over 64 has been clipped. (Log Scale, higher is better) )}}
  \label{exp-dgl-halfgnn-spmm-sddmmm}
\end{figure}

\subsubsection{\textbf{SpMM Runtime}}
DGL uses Cusparse for SpMMve. We measure the runtime for feature sizes 32 and 64. We compared our runtime against the Cusparse half-precision. Fig.~\ref{exp-dgl-halfgnn-spmm-sddmmm} shows the comparative speedup. We can observe that the speedup is massive for all datasets compared to Cusparse. Furthermore, these speedups are consistent over feature sizes 32 and 64. {\halfgnn} SpMMve achieves over $64\times$ runtime speedup in some cases, while achieving an average $22.89\times$ speedup. 

These speedups are observed for two reasons.
\textit{Firstly}, Cusparse half-precision SpMM introduces huge performance degradation compared to its own float-based SpMM as discussed earlier in Fig.~\ref{fig-float-half-spmm}. Indeed, our additional measurement shows that {\halfgnn} achieves a more realistic $2.52\times$ average speedup over Cusparse-float (not plotted). 
\textit{Secondly}, {\halfgnn} design (\cref{sec.optimization.spmm}) through its better data-load performance, increased computation throughput due to half2 usage, removal of atomic writes, workload balancing, etc., improves the {\halfgnn} kernel runtime. 


\begin{figure}[h]
  \centering
  \includegraphics[width=0.49\textwidth,]{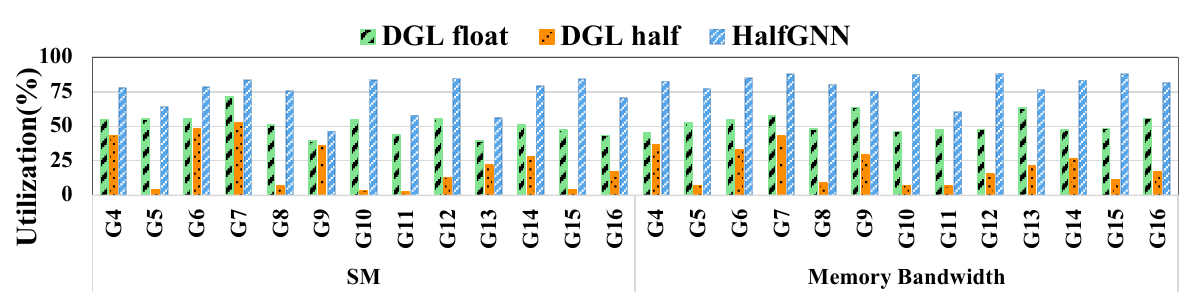} 
  \vspace{-17pt} 
  \caption{\small{{\halfgnn} SpMM achieves much higher SM and  memory bandwidth utilization compared to DGL float and DGL half }}
  \label{exp-SpMM-sm-memory}
 \vspace{-11pt} 
\end{figure}

To confirm, 
Fig.~\ref{exp-SpMM-sm-memory} plots a few performance counters using the Nvidia Nsight Compute (NCU) tool. 
{\halfgnn} achieves 80.92\% average \textit{utilization of memory bandwidth} compared to 20.22\% in Cusparse-half (which is used by DGL-half) and 51.99\% in Cusparse-float (which is used by DGL-float). The results show that {\halfgnn} has better data-load performance than DGL-half and DGL-float. 

\textit{Compute or SM utilization} for {\halfgnn} is 72.26\% compared to 21.58\% in Cusparse-half and 50.81\% in Cusparse-float. While the improvement is significant, it shows that the SpMM kernel is not compute-bound, a well-known fact. 
These design choices are explored again using micro-benchmarks in \cref{sec.exp.microbenchmark}. 
 
 SpMMv shares the same design principles and optimizations as SpMMve, whereas Cusparse does not provide this custom SpMM. Hence, we have not plotted it separately. 

\subsubsection{\textbf{SDDMM Runtime}}

\begin{figure}[t]
  \centering
  \includegraphics[width=0.45\textwidth,]{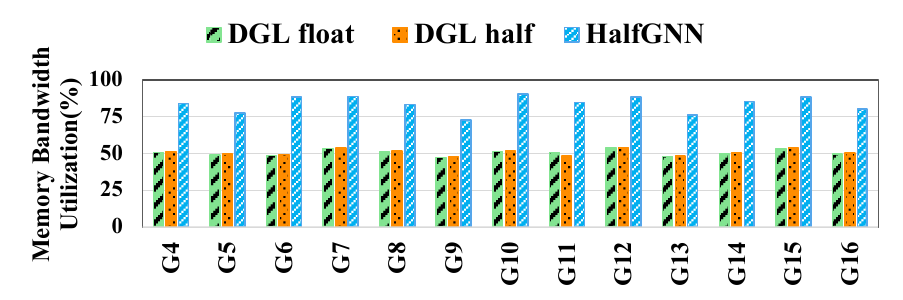} 
 \vspace{-11pt} 
  \caption{\small {\halfgnn} SDDMM achieves an average of 83.71\% memory bandwidth utilization compared to DGL float SDDMM (average of 50.59\%), and DGL half SDDMM (average of 50.85\%).}
  \label{exp-SDDMM-memory}
\end{figure}

Fig.~\ref{exp-dgl-halfgnn-spmm-sddmmm} also shows that  {\halfgnn} gets up to $12\times$ speedup, while most datasets get over $6\times$ speedup compared to the half-precision DGL \emph{SDDMM}. The average speedup is around $7.12\times$. 
The reason behind such a large speedup is that DGL \emph{half-precision} kernels individually do not provide any runtime advantage over their \emph{float} counterparts, as it just replaces float data-type with half without even using half2 or any other optimizations, as shown earlier in Fig.~\ref{fig-float-half-sddmmvv}. On the other hand, {\halfgnn} adopts half8 vector data-type which provides a huge performance boost in data-load time, and fewer rounds of inter-thread communication within a warp as discussed in \cref{sec.optimization.sddmm}. Nvidia's Compute Insight (NCU) tool confirms that {\halfgnn} achieves better memory bandwidth utilization of {83.71\%} compared to {50.85\%} in DGL-half and {50.59\%} in DGL-float as shown in Fig.~\ref{exp-SDDMM-memory}.

As DGL-half and DGL-float achieve similar memory bandwidth utilization, it empirically proves our hypothesis that merely replacing float with half would not provide any performance boost. 
Many of these design choices are evaluated separately in \cref{sec.exp.microbenchmark}.


\subsection{Micro-Benchmarking} \label{sec.exp.microbenchmark}
In this section, we perform many micro-benchmarks to measure the impact of a few proposed optimization choices.

\subsubsection{\textbf{SDDMM: Half8 vs Half2}}
\begin{figure}[h]
  \centering
  \includegraphics[scale=0.43]
  {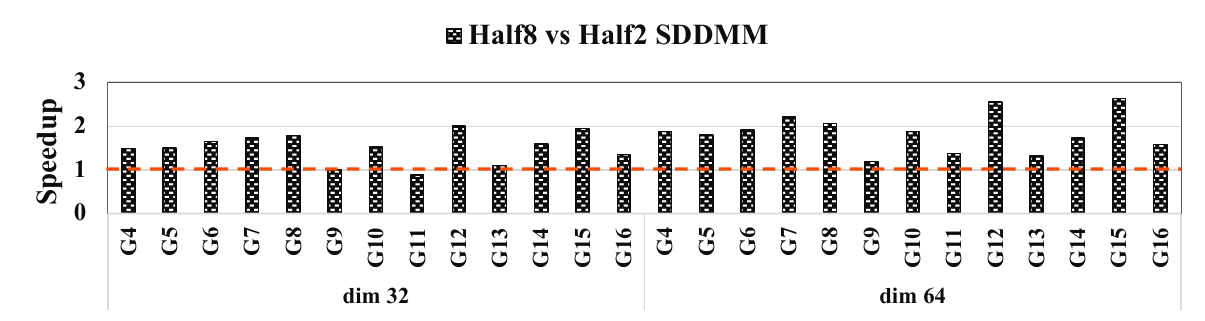} 
  \vspace{-11pt}
  \caption{\small Using \emph{half8} achieves $1.67\times$ speedup over \emph{half2} based SDDMM (Higher is better)}
  \label{exp-half8-vs-half2-sddmmvv-mul}
  \vspace{-11pt}
\end{figure}

We now show that usage of \textit{half8} is more efficient than using only \textit{half2} for \emph{SDDMM}. 
Fig.~\ref{exp-half8-vs-half2-sddmmvv-mul} shows that using \textit{half8} can provide runtime speedup up to almost $3\times$, and achieves an average speedup of $1.67\times$ for feature sizes $32$ and $64$. 
Using \textit{half8} reduces the number of required inter-thread communications for the reduction in \emph{SDDMM} and allows to issue load of four half2 features compared to just one.
This results in better data-load performance due to increased instruction-level parallelism as discussed in \cref{sec.optimization.sddmm}. 

\subsubsection{\textbf{SpMM: Atomic vs. Non-atomic Writes}} \label{sec.exp.atomic}


\begin{figure}[t]
  \centering
  \includegraphics[width=0.48\textwidth,]
  {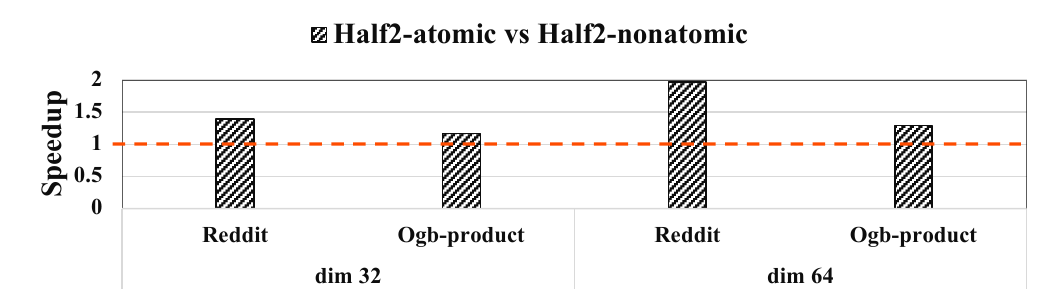} 
  \vspace{-11pt}
  \caption{\small {Speedup when atomic operations removed from {\halfgnn} SpMM}}
  \label{half2_atomic_vs_half2_nonatimc}
\end{figure}

Fig.~\ref {half2_atomic_vs_half2_nonatimc} shows the runtime speedup achieved when {\halfgnn} removes the atomic write through intra-CTA communication, staging buffer, and follow-up kernel compared to a version that uses atomic write, keeping everything as it is for edge-parallel design. This is true not only for Reddit and Ogb-product datasets, but also for all the other datasets.
Atomic implementations are more costly for \emph{half-precision} than float data-type. A large part of our design for \emph{SpMMve} revolved around how to get rid of atomic write (\cref{sec.optimization.spmm}). 

\subsubsection{\textbf{Huang-float vs Huang-Half2}} \label{sec.exp.huang}
We chose Huang et al.~\cite{huang2021understanding}, a state-of-the-art workload-balanced vertex-parallel work, to showcase the generality and effectiveness of our optimization proposed in {\halfgnn}.  We discussed the implementation ideas in \cref{sec.general} except that we did not change the neighbor grouping size and kept it to its original 32. Hence, a warp performs data loading edge features of 64 bytes and not 128 bytes due to batching of 32 neighbors. So, the new prototype compromises the memory coalescing. This is not a fundamental limitation, as we can change its neighbor group size to 64 to overcome the issue.




\begin{figure}[h]
  \centering
  \includegraphics[width=0.48\textwidth,]
  {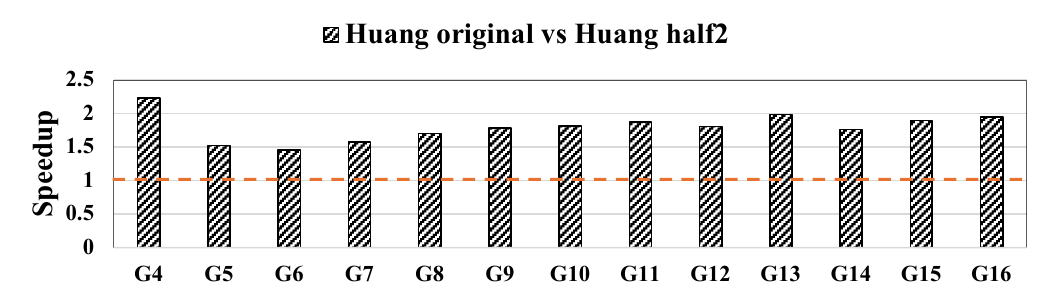} 
  \vspace{-18pt}
  \caption{\small {Huang-half2 gains $1.79\times$ speedup over Huang-float for its SpMM design.}}
  \label{exp-huang-float-vs-half2}
 \vspace{-4pt}
\end{figure}

Fig.~\ref{exp-huang-float-vs-half2} shows their runtime comparison, showing that the adopted half-precision version achieves around $1.79\times$ speedup on average. 
This is significant, as we achieve this without changing any fundamental nature of Huang et al., without applying even the full range of proposed optimizations. This demonstrates the generality of our proposed optimization. 
Although {\halfgnn} is an edge-parallel system, its design can also equally improve the performance of vertex-parallel systems. 

\section{Conclusion} \label{sec.conclusion}
We studied GNN training in half-precision, where we identified major issues that prevent training GNN efficiently and correctly. We started with building a GNN system using the half2 vector data type by overcoming several challenges. We then provided new vector data types and discretized reduction in half-precision training to achieve state-of-the-art system performance for SpMM, SDDMM, and overall GNN training and accuracy. 
Furthermore, our work provided a system that integrates the proposed techniques and has shown that the proposed design techniques apply to both the edge-parallel and the vertex-parallel systems. 


\section*{Acknowledgment}

We would like to thank the anonymous reviewer for their effort and feedback. The work is supported in part by National Science Foundation grant 2245849.  
The views, opinions, and findings presented in this material are those of the authors and do not represent the official views or
policies of the National Science Foundation.

\newpage



%






\nocite{*}

\balance
{

}
 
\end{document}